\documentclass{article}

\usepackage{natbib}

\usepackage[final]{neurips_2023}

\usepackage[utf8]{inputenc} 
\usepackage[T1]{fontenc}    
\usepackage{hyperref}       
\usepackage{url}            
\usepackage{booktabs}       
\usepackage{amsfonts}       
\usepackage{nicefrac}       
\usepackage{microtype}      
\usepackage{lmodern}

\usepackage{lmodern}
\usepackage[dvipsnames]{xcolor}
\usepackage{algpseudocode}
\usepackage{algorithm}
\usepackage{amsthm,amsfonts,amsmath,amssymb}
\usepackage{epsfig,color,float,graphicx,verbatim}
\usepackage{bbm}
\usepackage{subcaption}

\usepackage{array,ragged2e}
\newcolumntype{P}[1]{>{\RaggedRight\arraybackslash}p{#1}}

\usepackage{array}

\usepackage{hyperref}
\hypersetup{
	colorlinks   = true, 
	urlcolor     = blue, 
	linkcolor    = blue, 
	citecolor   = black 
}

\newtheorem{theorem}{Theorem}[section]

\newtheorem{corollary}{Corollary}[section]

\renewcommand{\eqref}[1]{Eq.~(\ref{#1})}

\newtheorem{open problem}[theorem]{Open Problem}

\usepackage{amssymb}

\usepackage{dsfont}

\newcommand{\stam}[1]{}

\usepackage{mathtools}

\newcommand{\bw}{\mathbf{w}}

\newcommand{\Ncal}{\mathcal{N}}

\newcommand{\reals}{{\mathbb R}}

\newcommand{\zero}{{\mathbf{0}}}

\newcommand{\inner}[1]{\langle #1 \rangle}
\newcommand{\norm}[1]{\left\|#1\right\|}

\title{Explaining high-dimensional text classifiers }

\author{%
  Odelia Melamed \\
  Microsoft R\&D, Israel\\
  \texttt{t-omelamed@microsoft.com} \\
  \And
  Rich Caruana \\
  Microsoft Research, Redmond\\
  \texttt{rcaruana@microsoft.com} \\
}

\begin{document}
\maketitle

\begin{abstract}
Explainability has become a valuable tool in the last few years, helping humans better understand AI-guided decisions. However, the classic explainability tools are sometimes quite limited when considering high-dimensional inputs and neural network classifiers. We present a new explainability method using theoretically proven high-dimensional properties in neural network classifiers. We present two usages of it: 1) On the classical sentiment analysis task for the IMDB reviews dataset, and 2) our Malware-Detection task for our PowerShell scripts dataset.

\end{abstract}

\section{Introduction}
In the last few years, Neural Networks have been commonly used for a variety of text-classification tasks (\citet{zhang2016rationale};
\citet{yang2016hierarchical}), from simple sentiment analysis to even fake news filtering (\citet{karmakharm2019journalist}). Unfortunately, those classifiers have trouble gaining human trust for two reasons: 1) The neural networks are considered black-box models, where one can see the inputs and output yet not understand the function in between; and 2) Textual inputs create non-continuous input space, which is hard to explore. A great effort has been devoted to developing methods to explain these models to increase trust and, therefore, their usability.
Several methods were set to overcome the black-box barrier, creating good explanations for predictions on continuous data domains (\citet{lundberg2017unified}, \citet{poerner2018evaluating}). However, this is only sometimes successful with continuous high-dimensional datasets (e.g., images). Experimentally, it has been shown that the well-known explainability tools on high-dimensional data and neural network classifiers behave differently.

If we look at it from the point of view of adversarial examples research, these inadequate explanations are yet another effect of the adversarial mystery (\citet{szegedy2013intriguing} and \citet{biggio2013evasion}). Experimentally, we witness this phenomenon in the case of neural network classifiers and high-dimensional data space. There, gradient calculations and other related methods experimentally lead us to adversarial examples: tiny, noise-looking changes in the input will mysteriously change the classifier decision (\citep{carlini2017adversarial, papernot2017practical, athalye2018obfuscated}). These small changes were experimentally shown as an out-of-distribution perturbation (\citet{shamir2021dimpled} and many others). Indeed, even in our security domain and coding inputs, adversarial examples have been studied as potential security hacks (\citet{schuster2021you}): one can change a minor part of the code input (that does not influence its execution) and change the classifier's prediction.

In the explainability research, some have noticed the connection between these out-of-distribution adversarial examples and the inadequate explanations for high-dimensional data (\citet{feng2018pathologies}). They note that the reason for the existence of these explanations is the high dimensionality of the input while implicitly relying on a lower-dimensional manifold (\citet{anders2020fairwashing}). Therefore, they suggest methods to ensure generating explanations within the data distribution (i.e., on the data manifold), using existing tools that are known to help avoid adversarial examples, such as Auto-Encoders (\citet{alvarez2017causal}), Generative models (\citet{chang2018explaining}), using surrogate models (\citet{anders2020fairwashing}), and others. 
Such data manifold exploration tools are not designed to be used when dealing with non-continuous input space such as text. Translating vector direction to a change in words is complex, so one cannot use simple vector manipulations. However, we can compromise for less - unlike adversarial robustness, detecting post-hoc the off-manifold examples is an easier task since no attacker is challenging the system.

In this paper, we are using a new perspective on adversarial examples to create informative explanations in the non-continuous input space. We use the theory from \citet{melamed2023adversarial} regarding neural networks trained on data relying on a low dimensional linear subspace to analyze the gradients off the data subspace. First, we note that the off-manifold gradients tend to have a big norm and, therefore, can be filtered using a simple threshold. In addition, we prove that different classifiers trained on the same data distribution will result in highly uncorrelated off-manifold gradients with high probability (using cosine similarity). We then use these conclusions in continuous input space regarding correlations and norms to create an "on-manifold" explanation in the non-continuous input space of text. 

Powershell is a common scripting language used by network administrators worldwide to perform anything from routine maintenance to complex admin tasks on a large number of machines. Unfortunately, attackers use these capabilities to develop malicious PowerShell scripts. As Powershell scripts are a widespread admin tool, we set out to classify malicious vs. benign scripts to detect malicious activity. As in many other domains, a Machine Learning based classification is accurate and frequently used. The explanation tools we provide are crucial for successful risk analysis and response. We utilize the many benefits of explanations, such as improved human trust, model and data evolution, and business intelligence. In addition, we can use predictions' explanations to identify the malicious areas in the code, mark them, and eventually prevent a comprehensive cyber-attack.

In our paper, we present a simple new method for creating on-manifold explanations. It will transfer easily between domains and even deployment environments. It only requires repeating the training for the same data distribution a few extra times, with no additional changes. This method is very approachable if the training set is very large or the training procedures are well-optimized and almost unchangeable, which is the case for many data-driven products today. We test it with two datasets: 1) the IMDB reviews dataset with the classical sentiment analysis task and 2) an industry Security PowerShell code dataset used to train models for malware detection. We present outstanding results in both settings compared to standard methods such as gradient max-norm, LIME, or SHAP.

\section{Related Work}
With the vast use of machine learning in text classification arises the demand for explanations of these classification decisions (\citet{ribeiro2016should}, \citet{wu2022survey}). 
For a few years now, scientists have been looking beyond correct classification to correct feature importance or silency maps (\citet{simonyan2013deep}, \citet{ross2017right}, \citet{ancona2017towards}, \citet{sundararajan2017axiomatic}). 
As many traditional explainability methods (\citet{lundberg2017unified}, \citet{poerner2018evaluating} and many others) frequently empirically failed to explain text classification tasks and particularly neural network classifiers, this subject has gained research interest.

In the last few years, researchers noticed some methods yield out-of-distribution explanations, sometimes referred to as explanations off the data manifold, and have tried to instead pursue on-manifold importance or explanations.
(\citet{wallace2018interpreting}) tried to explore the data manifold using the nearest neighbor. 
Some had manifold explorations solutions that apply to images or continuous data only (\citet{chang2018explaining},\citet{agarwal2020explaining},\citet{frye2020shapley},\citet{zhang2016rationale}) starting from encoding and decoding, explanations generators, and even more complicated models for explainability that require interaction (\citet{arous2021marta}).
Few trained robust models with some extra loss, sometimes require extra explanation information (\citet{ross2017right}, \citet{anders2020fairwashing},\citet{liu2018towards}).

In our paper, we start with a theoretical analysis of the problem and solutions and present a straightforward post-training explanation generation. Using no extra losses or special networks, which we believe is more accessible to the industry needs.
\section{Theoretical background}
\label{sec:theoretical}
In \citet{melamed2023adversarial}, the authors define the data distribution in a simplified way, as sampled from a low dimensional linear subspace of the input space. Using this simplification, they showed that the off-manifold gradients of the trained neural network have a large norm, hinting at the existence of nearby adversarial examples off the data distribution. The analysis of these gradients also yields a vital realization - they deviate very little from their initialization. 

In this section, we analyze the gradients according to these simplified settings. we denote the full input dimension as $d$, and the data subspace dimension as $d-\ell$. We denote an input sample by $x_0 \in \reals^d$, where $x_0 \sim M \subset \reals^d$ for some linear subspace $M$ of dimension $d-\ell$. Note that we cannot calculate $\ell$ exactly in real-life datasets. In general, when using simple methods for linear subspaces such as $PCA$, several off-manifold dimensions can be approximated. Yet, especially in cases when the data manifold is not exactly linear, other off-manifold dimensions might be accidentally considered as on-manifold dimensions. 

\subsection{Gradient Norms}
In \citet{melamed2023adversarial}, the only restriction on the dataset is to lie on $M$, with no constraints on the distance between the data points. Surprisingly, under reasonable assumptions, the authors prove a lower bound of the off manifold gradient norm of $\Omega(1)$, which hints at a close-by adversarial example (i.e., an example from the opposite class). 

\subsection{Cosine Similarity between Same-Input-Gradients}
It was shown in \citet{melamed2023adversarial} that the off-manifold gradients are affected mainly by the initialization. Therefore, when we train two different neural networks $N_1$, $N_2$ with the same training method on the same training data distribution, one can expect that the two off-manifold gradients would be very non-correlated (with cosine-similarity of some sub-exponential with  $\approx \frac{1}{\ell}$ variance). 
Formally, we use the simplified setting of a two-layer, fully connected neural network to show exponential concentration bound for the inner product between the gradients. We define:
\[
N_1(x,\bw^1_{1:m})  = \sum\limits_{i=1}^{m} u_i \sigma({w_i^1}^\top x),\text{    } N_2(x,\bw^2_{1:m})  = \sum\limits_{i=1}^{m} v_i \sigma({w_i^2}^\top x)~.
\]
\begin{theorem}\label{thm:innerbounded}
Let an input sample $x_0 \in M \subset \reals^d$. For neural network $N_1$, let $S_1= \{i\in[m]: \inner{w^1_i,x_0} \geq 0\}$, and let $k_1 := |S_1|$. Similarly for $N_2$, $S_2= \{i\in[m]: \inner{w^2_i,x_0} \geq 0\}$, and let $k_2 := |S_2|$. We denote by $g_i$ the gradient of the network $N_i$ with respect to the input at $x_0$, i.e. $g_i = \frac{\partial N_i(x_0)}{\partial x}$, and its projection on $M$ by $\Tilde{g}_i$ i.e. $\Tilde{g}_i = \Pi_{M^\perp}\left(\frac{\partial N_i(x_0)}{\partial x}\right)$. Then:
\[
\Pr \left[|\inner{\Tilde{g}_1, \Tilde{g}_2}|  \geq \frac{\sqrt{2\ell}}{d}    \right] \leq  e^{-\ell/16} + 2e^{-m/2}~.
\]
\end{theorem}
The full proof can be found in appendix \ref{appen:proof}. In short, we note that:
\[
     \Tilde{g}_1 =  \sum\limits_{i \in S_1} u_i \hat{w}^1_i \text{   ,   } \Tilde{g}_2 =  \sum\limits_{i \in S_2} v_i \hat{w}^2_i 
\] 
and therefore,
\begin{align*}
    |\inner{\Tilde{g}_1, \Tilde{g}_2}| &= |\inner{\sum\limits_{i \in S_1} u_i \hat{w}^1_i , \sum\limits_{i \in S_2} v_i \hat{w}^2_i}| = \frac{1}{m^2}|\inner{\sum\limits_{i \in S_1} sign\left(u_i\right) \hat{w}^1_i , \sum\limits_{i \in S_2} sign\left(v_i\right) \hat{w}^2_i}|~.
    \end{align*}
We also note that $\sum\limits_{i \in S_1} sign\left(u_i\right) \hat{w}^1_i \sim \Ncal\left(\zero, \frac{k_1}{d}I_\ell\right) $, and $ \sum\limits_{i \in S_2} sign\left(v_i\right) \hat{w}^2_i \sim \Ncal\left(\zero, \frac{k_2}{d}I_\ell\right)$. We then use concentration bounds provided in the paper to conclude the proof.
\begin{corollary}\label{cor:innerbounded}
In the settings of Theorem \ref{thm:innerbounded}, assume that $\ell = \Theta(d)$ and $k = \Theta(m)$. Then, with probability $\geq 1 - (2e^{-\Theta(d)} + 2e^{- m})$:
\[
|S_C(\Tilde{g}_1, \Tilde{g}_2)|  \leq \Theta\left(\frac{1}{\sqrt{\ell}}\right)~.
\]

\end{corollary}
See Appendix \ref{appen:proof} for details. As expected, under the original assumptions, we showed that the gradient vectors w.r.t. the same input for two differently initialized networks are highly non-correlated, by giving an upper bound for their cosine similarity.
\section{Our Method - Theory to Practice}
In text, a pre-processed and embedded input sample is a 2D matrix $x_0 \in \mathrm{R}^n \times \mathrm{R}^p$, where $n$ is the number of words in the input (padded or clipped if needed), and $p$ is the embedding dimension chosen for each word. One main limitation exists when using gradient-based tools on a textual dataset rather than continuous data - we cannot explore the input space by changing an input word in the direction of the gradient. Therefore, we usually look at each word's gradient norm to determine its significance rather than its direction. 
In this section, we use the theoretical observations to find the "on-manifold" gradients in settings where the data manifold is not easily found or even defined (e.g., in the case of a non-continuous data manifold). 

Separating the input gradient into $n$ words' gradients of dimension $p$, we wish to understand if a word's gradient is mostly on or off the data manifold. Let $C$ be the classifier we are explaining. We denote by $g_C$ the gradient of $C$ with respect to the input at $x_0$ (i.e., $g_C = \frac{\partial C(x_0)}{\partial x}$). Note that $g_C \in \mathrm{R}^n \times \mathrm{R}^p$. We denote by $g_C^j$ the gradient of $C$ with respect to the $j$-th word of the input at $x_0$. We say that the word's gradient is mostly off-manifold if most of its coordinates are off-manifold coordinates. We wish to detect the gradients that are mostly off-manifold and keep the ones that are mostly on the manifold.

\subsection{Expected Gradient Norms}
We assume that our data approximately lie within a low dimensional linear subspace. In addition, assuming the inputs lie within a $O(\sqrt{d})$ distance from each other is standard in real-life data. In this case, one can expect an average gradient of $O(\frac{1}{\sqrt{d}})$ along the shortest path between two different input samples (this path also lies in the linear subspace). It is easy to see that an $\Omega(1)$ norm of the off manifold gradient is very surprising. Consequently, one can expect the off-manifold gradient to have a relatively large norm, also when divided into $p$-dimensional vectors. Specifically, for any $j \in [n]$, we look at $\norm{g_C^j}$. 

\subsection{Expected Cosine Similarity}
Let $\{N_i\}_{i=1}^t$ be our surrogate classifiers ensemble, where each classifier has been trained on the same training distribution as $C$ with different initialization weights. We denote by $g_i$ the gradient of the surrogate network $N_i$ with respect to the input at $x_0$, i.e. $g_i = \frac{\partial N_i(x_0)}{\partial x}$. Note, $g_i \in \mathrm{R}^n \times \mathrm{R}^p$. Similarly to $g_C^j$, we denote $g_i^j$ as the gradient of $N_i$ with respect to the $j$-th word of the input at $x_0$. Note, $g_C^j, g_i^j \in \mathrm{R}^p$. Now, for any $j \in [n]$, we look at $\alpha_{g_C^j} = \frac{1}{t} \sum\limits_{i =0}^{t} |S_C(g_C^j, g_i^j)|$. 

\subsection{The Method}
Our method aims to find the words' gradients that are mostly on-manifold. If the gradient was mostly off-manifold, we expect it to have a relatively large norm and slight cosine similarity with the corresponding gradients of the other networks, i.e., small $\alpha_{g_C^j}$ (as explained in \ref{sec:theoretical}). Therefore, looking for the on-manifold gradient, we take the maximal $\alpha_{g_C^j}$, between those with relatively small gradient norms. 
The relevant norm threshold for each dataset should be of size $O(\frac{1}{\sqrt{p}})$. Since the $O$ notation can hide any constant, one should test the dataset's gradient's norms distribution to easily understand the relevant norm threshold, see Section \ref{section:imdb} and Appendix \ref{appen:imdb} and \ref{appen:Security} for examples.

\begin{algorithm}
\caption{Choose k top words for explanation}\label{alg:cap}

\begin{algorithmic}[1]
\Require Classifier $C$, surrogate classifiers $\{N_i\}_{i=1}^t$, norm treshold $T$
\State $g_C \gets \frac{\partial C(x_0)}{\partial x} \in \mathrm{R}^n \times \mathrm{R}^p$, 
\State $\forall i \in [t]$, $g_i \gets \frac{\partial N_i(x_0)}{\partial x} \in \mathrm{R}^n \times \mathrm{R}^p$ 
\State $small-norms \gets \{j : \norm{g_C^j} < T\}  $
\State $avarage-cosine-similarities \gets \{ \alpha_{g_C^j} : j \in  small-norms\}$
\State \Return $\text{Max-k} \left( avarage-cosine-similarities \right)$

\end{algorithmic}
\end{algorithm}
\section{Experiment - IMDB dataset}\label{section:imdb}
\subsection{Data Manifold analysis Results}
\begin{figure}[h]
\includegraphics[width=6cm]{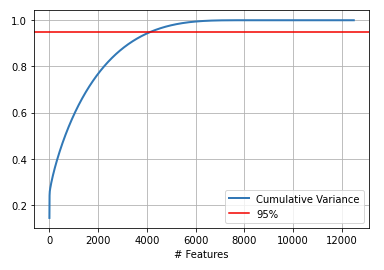}
  \centering
  \caption{PCA for IMDB Sentiment Analysis Dataset. Showing cumulative variance for input dimension $d=32,000$ we can see that the first $4000$ already gets 95\% of the variance, and the first $6000$ are enough. So, here $d-l \leq 6000$. }
  \label{img:imdb_pca}
\end{figure}

We use a Sentiment Analysis classifier for  IMDB reviews. On this dataset, the input dimension is $d=32,000$, $n=500$ words are taken for each review, and the embedding size is $p=64$. 
Figure \ref {img:imdb_pca} shows the implicit low-dimensional linear subspace for this dataset using PCA decomposition. One can see that we can separate many "zero"-data dimensions (zero up to a rotation) from the data subspace. The accumulated variance reaches $1$ after considering only the first $6000$ features. 
Note that for the explanation we disregard the padding, so there are fewer off-manifold dimensions expected for the text only. This figure hints that the separation between off and on-manifold gradients will be beneficial. 

\subsection{Choosing norm threshold}
\begin{figure}[H]
\includegraphics[width=0.7\textwidth]{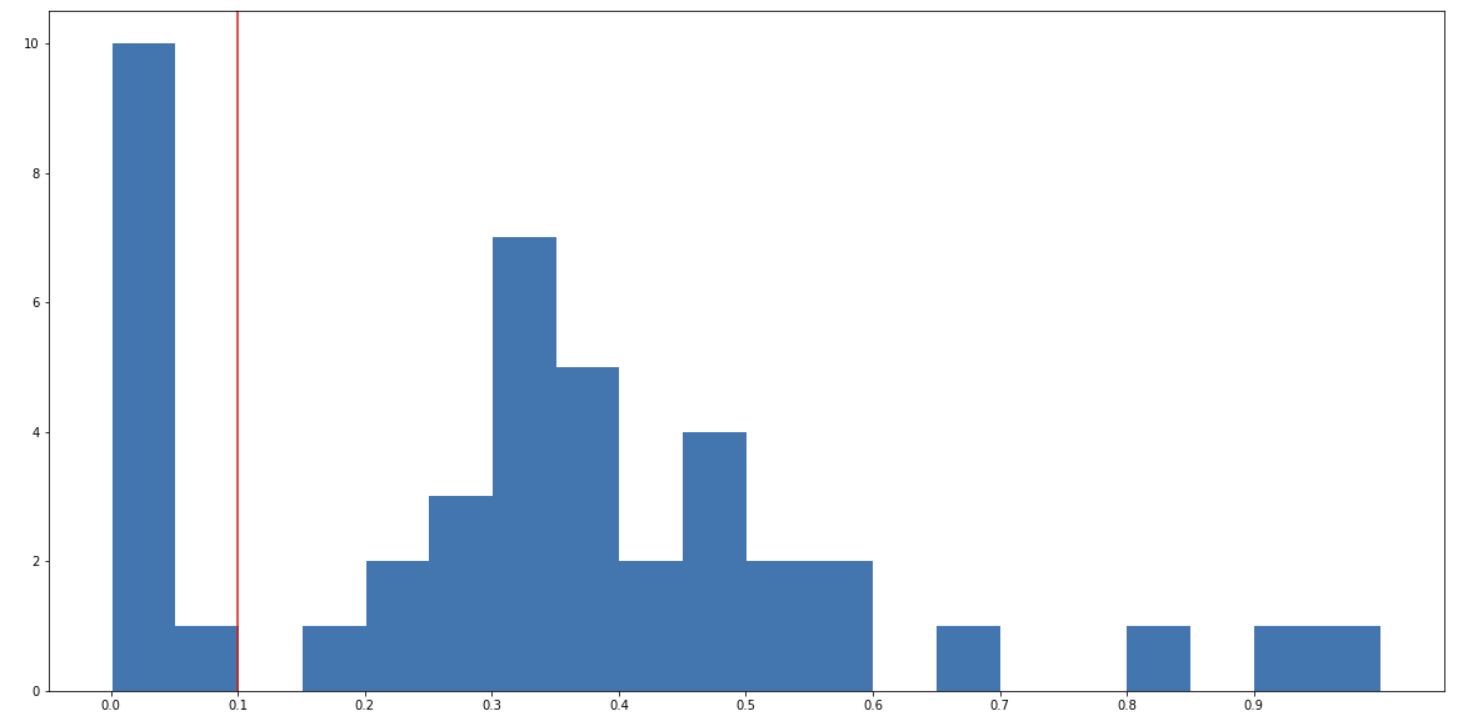}
  \centering
  \caption{Word Gradient Norm for IMDB Sentiment Analysis Dataset. One can see a clear gap around $0.1$.}
  \label{img:imdb_norms}
  
\end{figure}

In Figure \ref{img:imdb_norms} we plot in a histogram the different norms, normalized using $L_{\infty}$ norm for convenience, of the words in the input sentence. In the figure, one can see a clear gap between words above and below the threshold of $0.1$. The histogram shows 1) many words with norms smaller than $0.1$, 2) most norms in the middle section (which makes sense after observing many off-manifold dimensions in the previous section), and 3) a few with larger norms. For visualization, the negligible norms (less than $e^{-3}$) are filtered. Altogether, This histogram hint on $0.1$ as a good candidate for norm threshold for this dataset. This simple test can be done for each input separately or for the entire dataset together, according to user preferences and application.

\subsection{Explanation Results}
After choosing a norm threshold for this dataset, we take the top ten words $j$ with the maximal $\alpha_{g_C^j}$ among those with $\norm{g_C^j}$ smaller than the norm threshold. Appendix \ref{appen:imdb} provides more details about this experiment. Table \ref{tbl_imdb} compares our new explanation method to the classical gradient-based method of taking top-norm words. One can see that the top norm words do not relate to negative sentiment and are neutral even in context. Similar explanations of this text using LIME and SHAP methods can be found in Appendix \ref{appen:imdb}. The words picked by our method are clearly related to negative sentiment, e.g., "poor story", "impossible plot", etc..
\begin{table}[h]
  \caption{IMDB Sentiment Explanation}
  \label{imxb-example-table}
  \centering
    \newlength\qd
\setlength\qd{\dimexpr .91\textwidth -2\tabcolsep}
  \newlength\ql
\setlength\ql{\dimexpr .09\textwidth -2\tabcolsep}
  \begin{tabular}{{P{\ql}P{\qd}}}
    \toprule
    \midrule
By Norm & \scriptsize Seems Sensei Seagal is getting more and more moralising and less "action packed". To date this has to be his worse movie... no action, a poor story line, an impossible plot and to make things worse, one of the CHEEZIEST endings I have ever seen. Seagal films are like \textcolor{BurntOrange}{seeing} a Dirty-Harry, you do not \textcolor{BurntOrange}{go} \textcolor{BurntOrange}{see} it for the \textcolor{BurntOrange}{great} \textcolor{BurntOrange}{social} causes or impeccable acting... you \textcolor{BurntOrange}{want} a \textcolor{BurntOrange}{good} \textcolor{BurntOrange}{action} flick. On a scale of 1 to 10, this \textcolor{BurntOrange}{one} \textcolor{BurntOrange}{gets} a 1... \\
Ours &  \scriptsize Seems Sensei Seagal is getting more and more moralising and less "action packed". To date this has to be his \textcolor{BurntOrange}{worse} movie... no action, a \textcolor{BurntOrange}{poor} \textcolor{BurntOrange}{story} line, an \textcolor{BurntOrange}{impossible} \textcolor{BurntOrange}{plot} and to \textcolor{BurntOrange}{make} \textcolor{BurntOrange}{things} worse, \textcolor{BurntOrange}{one} of the CHEEZIEST endings I have \textcolor{BurntOrange}{ever} seen. Seagal \textcolor{BurntOrange}{films} are like seeing a Dirty-Harry, you do not go see it for the great social causes or impeccable acting... you want a good action flick. On a scale of 1 to 10, this one gets a 1... \\
    \bottomrule
  \end{tabular}
  \label{tbl_imdb}
  \caption{IMDB movie review labeled \textbf{negative} and Neural Network sentiment model explained by vector norm and by our method. Orange words are the top 10 for each method.}
\end{table}

\section{Security Dataset Experiment}\label{sec:security}
In the Security context, explainability tools can be a real game-changer for the malware detection task. As millions of lines of scripts are executed hourly on each computer in an organization, it's very hard to detect malware spreading and threatening the organization's computers and data. In addition, this classification task is constantly challenged by hackers. Potential attackers keep trying to manipulate their code to avoid getting caught by the detector while maintaining its harmful functionality. Therefore, trusting the detector to identify the harmful parts correctly is particularly crucial.
For a simple example, we look at the code: 
\[ \textbf{"foreach (\$harmful\_item in \$harmful\_set)
\{<harmful functionality>\}".}\]

If the detector recognizes malware-related activity thanks to the variable names "\$harmful\_item" and "\$harmful\_set", the attacker will simply change their names and bypass our detector (\citet{schuster2021you}). We want to ensure that the detector can identify harmful functionality given different variable names. Note that creating the correct explanation does not mean our detector is not still sensitive to adversarial examples, but a correct explanation will help us to adjust its functionality when it is mistaken. 

\subsection{Explanation Results}
For this experiment, we chose the same norm threshold for this dataset as in the IMDB experiment. Similarly, we take the top ten words $j$ with the maximal $\alpha_{g_C^j}$ among those with $\norm{g_C^j}$ smaller than the norm threshold. More about the experiment details and threshold choice are in Appendix \ref{appen:Security}.

Table \ref{tbl_PowerShell} shows PowerShell code detected as malware, explained by the classical gradient-based by word norm, and using our method. The norm-based method mistakenly prefers the first few tokens that are just general security networking setups. In Appendix \ref{appen:Security}, we show similar explanations using the LIME and SHAP algorithms. At the bottom, we show our method chooses several important expressions commonly related to malware: the "URI" is mandatory for functionality and commonly used to download malicious content from a remote server. "Convert", "FromBase64String", and "UTF8" terms are needed to make downloaded content into an executable script, and the "Invoke-Expression" is mandatory to execute that script, so this is an excellent explanation.
%
%
\begin{table}[H]
  \caption{PowerShell Detection Explanation}
  \centering
  \newlength\qf
\setlength\qf{\dimexpr .91\textwidth -2\tabcolsep}
  \newlength\qg
\setlength\qg{\dimexpr .09\textwidth -2\tabcolsep}
  \begin{tabular}{{P{\qg}P{\qf}}}
    \toprule
    \midrule

  By Norm & \scriptsize [\textcolor{BurntOrange}{Net}.\textcolor{BurntOrange}{ServicePointManager}]::\textcolor{BurntOrange}{SecurityProtocol} = [\textcolor{BurntOrange}{Net}.\textcolor{BurntOrange}{SecurityProtocolType}]::\textcolor{BurntOrange}{Tls12};\$xor = [\textcolor{BurntOrange}{System}.\textcolor{BurntOrange}{Text}.Encoding]::UTF8.GetBytes('**-**-**');\$base64String = (Invoke-WebRequest -URI https://**.\textcolor{BurntOrange}{blob}.\textcolor{BurntOrange}{core}.**.**/**/**.txt -UseBasicParsing).Content;Try\{ \$contentBytes = [System.Convert]::FromBase64String(\$base64String) \} Catch \{ \$contentBytes = [System.Convert]::FromBase64String(\$base64String.Substring(3)) \};\$i = 0; \$ decryptedBytes = @();\$contentBytes.foreach\{ \$decryptedBytes += \$ \_ -bxor \$xor[\$i]; \$i++; if (\$i -eq \$xor.Length) \{ \$i = 0\} \};Invoke-Expression ([System.Text.Encoding]::UTF8.GetString(\$decryptedBytes)) \\
 
 Ours & \scriptsize [Net.ServicePointManager]::SecurityProtocol = [Net.SecurityProtocolType]::Tls12;\$xor = [System.Text.Encoding]::UTF8.GetBytes('**-**-**');\$base64String = (Invoke-WebRequest \textcolor{BurntOrange}{-URI} https://**.blob.core.**.**/**/**.txt -UseBasicParsing).Content;Try\{ \$contentBytes = [System.\textcolor{BurntOrange}{Convert}]::FromBase64String(\$base64String) \} Catch \{ \$contentBytes = [System.Convert]::\textcolor{BurntOrange}{FromBase64String}(\$base64String.Substring(3)) \};\$i = 0; \$ decryptedBytes = @();\$contentBytes.foreach\{ \$decryptedBytes += \$ \_ -bxor \$xor[\textcolor{BurntOrange}{\$i}]; \$i++; \textcolor{BurntOrange}{if} (\$i \textcolor{BurntOrange}{-eq} \$xor.\textcolor{BurntOrange}{Length}) \{ \textcolor{BurntOrange}{\$i} = 0\} \};\textcolor{BurntOrange}{Invoke-Expression} ([System.Text.Encoding]::\textcolor{BurntOrange}{UTF8}.GetString(\$decryptedBytes)) \\
    \bottomrule
  \end{tabular}
  \caption{PowerShell script labeled \textbf{malware-related} and Neural Network classifier explained by vector norm and by our method. Orange-colored words are the top 10 for each method.}
  \label{tbl_PowerShell}
\end{table}

\section{Conclusions and Future Work}\label{sec:conclusions}
In this paper we presented a novel method for creating on-manifold explanations, using a recent theoretical model from adversarial examples research. We presented a natural language use of this method on the IMDB sentiment analysis task, as well as on industrial scripts dataset for the malware detection task.
One interesting research area using our theoretical work is understanding the gradient behaviour for implicit low-dimensional datasets for network architectures that are used in text-classification tasks.
Another research area we find inspiring is the interdisciplinary point of view, i.e., the on manifold exploration in general. Currently, there is an extensive research effort to approximate and explore the data manifold, and the explainability could benefit from this research. One can use these efforts to help generate on-manifold post-hoc explanations in many different settings. In the other direction, An interesting future direction is how to enhance adversarial robustness using tips from explanations.

\newpage
\bibliographystyle{plainnat}
\bibliography{main}

\begin{thebibliography}{28}
\providecommand{\natexlab}[1]{#1}
\providecommand{\url}[1]{\texttt{#1}}
\expandafter\ifx\csname urlstyle\endcsname\relax
  \providecommand{\doi}[1]{doi: #1}\else
  \providecommand{\doi}{doi: \begingroup \urlstyle{rm}\Url}\fi

\bibitem[Agarwal and Nguyen(2020)]{agarwal2020explaining}
Chirag Agarwal and Anh Nguyen.
\newblock Explaining image classifiers by removing input features using generative models.
\newblock In \emph{Proceedings of the Asian Conference on Computer Vision}, 2020.

\bibitem[Alvarez-Melis and Jaakkola(2017)]{alvarez2017causal}
David Alvarez-Melis and Tommi~S Jaakkola.
\newblock A causal framework for explaining the predictions of black-box sequence-to-sequence models.
\newblock \emph{arXiv preprint arXiv:1707.01943}, 2017.

\bibitem[Ancona et~al.(2017)Ancona, Ceolini, {\"O}ztireli, and Gross]{ancona2017towards}
Marco Ancona, Enea Ceolini, Cengiz {\"O}ztireli, and Markus Gross.
\newblock Towards better understanding of gradient-based attribution methods for deep neural networks.
\newblock \emph{arXiv preprint arXiv:1711.06104}, 2017.

\bibitem[Anders et~al.(2020)Anders, Pasliev, Dombrowski, M{\"u}ller, and Kessel]{anders2020fairwashing}
Christopher Anders, Plamen Pasliev, Ann-Kathrin Dombrowski, Klaus-Robert M{\"u}ller, and Pan Kessel.
\newblock Fairwashing explanations with off-manifold detergent.
\newblock In \emph{International Conference on Machine Learning}, pages 314--323. PMLR, 2020.

\bibitem[Arous et~al.(2021)Arous, Dolamic, Yang, Bhardwaj, Cuccu, and Cudr{\'e}-Mauroux]{arous2021marta}
Ines Arous, Ljiljana Dolamic, Jie Yang, Akansha Bhardwaj, Giuseppe Cuccu, and Philippe Cudr{\'e}-Mauroux.
\newblock Marta: Leveraging human rationales for explainable text classification.
\newblock In \emph{Proceedings of the AAAI conference on artificial intelligence}, volume~35, pages 5868--5876, 2021.

\bibitem[Athalye et~al.(2018)Athalye, Carlini, and Wagner]{athalye2018obfuscated}
Anish Athalye, Nicholas Carlini, and David Wagner.
\newblock Obfuscated gradients give a false sense of security: Circumventing defenses to adversarial examples.
\newblock In \emph{International conference on machine learning}, pages 274--283. PMLR, 2018.

\bibitem[Biggio et~al.(2013)Biggio, Corona, Maiorca, Nelson, {\v{S}}rndi{\'c}, Laskov, Giacinto, and Roli]{biggio2013evasion}
Battista Biggio, Igino Corona, Davide Maiorca, Blaine Nelson, Nedim {\v{S}}rndi{\'c}, Pavel Laskov, Giorgio Giacinto, and Fabio Roli.
\newblock Evasion attacks against machine learning at test time.
\newblock In \emph{Joint European conference on machine learning and knowledge discovery in databases}, pages 387--402. Springer, 2013.

\bibitem[Carlini and Wagner(2017)]{carlini2017adversarial}
Nicholas Carlini and David Wagner.
\newblock Adversarial examples are not easily detected: Bypassing ten detection methods.
\newblock In \emph{Proceedings of the 10th ACM workshop on artificial intelligence and security}, pages 3--14, 2017.

\bibitem[Chang et~al.(2018)Chang, Creager, Goldenberg, and Duvenaud]{chang2018explaining}
Chun-Hao Chang, Elliot Creager, Anna Goldenberg, and David Duvenaud.
\newblock Explaining image classifiers by counterfactual generation.
\newblock \emph{arXiv preprint arXiv:1807.08024}, 2018.

\bibitem[Feng et~al.(2018)Feng, Wallace, Grissom~II, Iyyer, Rodriguez, and Boyd-Graber]{feng2018pathologies}
Shi Feng, Eric Wallace, Alvin Grissom~II, Mohit Iyyer, Pedro Rodriguez, and Jordan Boyd-Graber.
\newblock Pathologies of neural models make interpretations difficult.
\newblock \emph{arXiv preprint arXiv:1804.07781}, 2018.

\bibitem[Frye et~al.(2020)Frye, de~Mijolla, Begley, Cowton, Stanley, and Feige]{frye2020shapley}
Christopher Frye, Damien de~Mijolla, Tom Begley, Laurence Cowton, Megan Stanley, and Ilya Feige.
\newblock Shapley explainability on the data manifold.
\newblock \emph{arXiv preprint arXiv:2006.01272}, 2020.

\bibitem[Karmakharm et~al.(2019)Karmakharm, Aletras, and Bontcheva]{karmakharm2019journalist}
Twin Karmakharm, Nikolaos Aletras, and Kalina Bontcheva.
\newblock Journalist-in-the-loop: Continuous learning as a service for rumour analysis.
\newblock In \emph{Proceedings of the 2019 Conference on Empirical Methods in Natural Language Processing and the 9th International Joint Conference on Natural Language Processing (EMNLP-IJCNLP): System Demonstrations}, pages 115--120, 2019.

\bibitem[Liu et~al.(2018)Liu, Yin, and Wang]{liu2018towards}
Hui Liu, Qingyu Yin, and William~Yang Wang.
\newblock Towards explainable nlp: A generative explanation framework for text classification.
\newblock \emph{arXiv preprint arXiv:1811.00196}, 2018.

\bibitem[Lundberg and Lee(2017)]{lundberg2017unified}
Scott~M Lundberg and Su-In Lee.
\newblock A unified approach to interpreting model predictions.
\newblock \emph{Advances in neural information processing systems}, 30, 2017.

\bibitem[Melamed et~al.(2023)Melamed, Yehudai, and Vardi]{melamed2023adversarial}
Odelia Melamed, Gilad Yehudai, and Gal Vardi.
\newblock Adversarial examples exist in two-layer relu networks for low dimensional data manifolds.
\newblock \emph{arXiv preprint arXiv:2303.00783}, 2023.

\bibitem[Papernot et~al.(2017)Papernot, McDaniel, Goodfellow, Jha, Celik, and Swami]{papernot2017practical}
Nicolas Papernot, Patrick McDaniel, Ian Goodfellow, Somesh Jha, Z~Berkay Celik, and Ananthram Swami.
\newblock Practical black-box attacks against machine learning.
\newblock In \emph{Proceedings of the 2017 ACM on Asia conference on computer and communications security}, pages 506--519, 2017.

\bibitem[Poerner et~al.(2018)Poerner, Roth, and Sch{\"u}tze]{poerner2018evaluating}
Nina Poerner, Benjamin Roth, and Hinrich Sch{\"u}tze.
\newblock Evaluating neural network explanation methods using hybrid documents and morphological agreement.
\newblock \emph{arXiv preprint arXiv:1801.06422}, 2018.

\bibitem[Ribeiro et~al.(2016)Ribeiro, Singh, and Guestrin]{ribeiro2016should}
Marco~Tulio Ribeiro, Sameer Singh, and Carlos Guestrin.
\newblock " why should i trust you?" explaining the predictions of any classifier.
\newblock In \emph{Proceedings of the 22nd ACM SIGKDD international conference on knowledge discovery and data mining}, pages 1135--1144, 2016.

\bibitem[Ross et~al.(2017)Ross, Hughes, and Doshi-Velez]{ross2017right}
Andrew~Slavin Ross, Michael~C Hughes, and Finale Doshi-Velez.
\newblock Right for the right reasons: Training differentiable models by constraining their explanations.
\newblock \emph{arXiv preprint arXiv:1703.03717}, 2017.

\bibitem[Schuster et~al.(2021)Schuster, Song, Tromer, and Shmatikov]{schuster2021you}
Roei Schuster, Congzheng Song, Eran Tromer, and Vitaly Shmatikov.
\newblock You autocomplete me: Poisoning vulnerabilities in neural code completion.
\newblock In \emph{30th USENIX Security Symposium (USENIX Security 21)}, pages 1559--1575, 2021.

\bibitem[Shamir et~al.(2021)Shamir, Melamed, and BenShmuel]{shamir2021dimpled}
Adi Shamir, Odelia Melamed, and Oriel BenShmuel.
\newblock The dimpled manifold model of adversarial examples in machine learning.
\newblock \emph{arXiv preprint arXiv:2106.10151}, 2021.

\bibitem[Simonyan et~al.(2013)Simonyan, Vedaldi, and Zisserman]{simonyan2013deep}
Karen Simonyan, Andrea Vedaldi, and Andrew Zisserman.
\newblock Deep inside convolutional networks: Visualising image classification models and saliency maps.
\newblock \emph{arXiv preprint arXiv:1312.6034}, 2013.

\bibitem[Sundararajan et~al.(2017)Sundararajan, Taly, and Yan]{sundararajan2017axiomatic}
Mukund Sundararajan, Ankur Taly, and Qiqi Yan.
\newblock Axiomatic attribution for deep networks.
\newblock In \emph{International conference on machine learning}, pages 3319--3328. PMLR, 2017.

\bibitem[Szegedy et~al.(2013)Szegedy, Zaremba, Sutskever, Bruna, Erhan, Goodfellow, and Fergus]{szegedy2013intriguing}
Christian Szegedy, Wojciech Zaremba, Ilya Sutskever, Joan Bruna, Dumitru Erhan, Ian Goodfellow, and Rob Fergus.
\newblock Intriguing properties of neural networks.
\newblock \emph{Preprint, arXiv:1312.6199}, 2013.

\bibitem[Wallace et~al.(2018)Wallace, Feng, and Boyd-Graber]{wallace2018interpreting}
Eric Wallace, Shi Feng, and Jordan Boyd-Graber.
\newblock Interpreting neural networks with nearest neighbors.
\newblock \emph{arXiv preprint arXiv:1809.02847}, 2018.

\bibitem[Wu et~al.(2022)Wu, Xiao, Sun, Zhang, Ma, and He]{wu2022survey}
Xingjiao Wu, Luwei Xiao, Yixuan Sun, Junhang Zhang, Tianlong Ma, and Liang He.
\newblock A survey of human-in-the-loop for machine learning.
\newblock \emph{Future Generation Computer Systems}, 135:\penalty0 364--381, 2022.

\bibitem[Yang et~al.(2016)Yang, Yang, Dyer, He, Smola, and Hovy]{yang2016hierarchical}
Zichao Yang, Diyi Yang, Chris Dyer, Xiaodong He, Alex Smola, and Eduard Hovy.
\newblock Hierarchical attention networks for document classification.
\newblock In \emph{Proceedings of the 2016 conference of the North American chapter of the association for computational linguistics: human language technologies}, pages 1480--1489, 2016.

\bibitem[Zhang et~al.(2016)Zhang, Marshall, and Wallace]{zhang2016rationale}
Ye~Zhang, Iain Marshall, and Byron~C Wallace.
\newblock Rationale-augmented convolutional neural networks for text classification.
\newblock In \emph{Proceedings of the Conference on Empirical Methods in Natural Language Processing. Conference on Empirical Methods in Natural Language Processing}, volume 2016, page 795. NIH Public Access, 2016.

\end{thebibliography}

\newpage
\appendix
\section{Proof from Section \ref{sec:theoretical}}\label{appen:proof}
\begin{proof}
we noted that:

\[
     \Pi_{M^\perp}\left(\frac{\partial N_1(x_0)}{\partial x}\right) =  \sum\limits_{i \in S_1} u_i \hat{w}^1_i \text{   ,   } \Pi_{M^\perp}\left(\frac{\partial N_2(x_0)}{\partial x}\right) =  \sum\limits_{i \in S_2} v_i \hat{w}^2_i ~.
\] 
Therefore,

\begin{align*}
    |\inner{\Tilde{g}_1, \Tilde{g}_2}| &= |\inner{\sum\limits_{i \in S_1} u_i \hat{w}^1_i , \sum\limits_{i \in S_2} v_i \hat{w}^2_i}| = \frac{1}{m^2}|\inner{\sum\limits_{i \in S_1} sign\left(u_i\right) \hat{w}^1_i , \sum\limits_{i \in S_2} sign\left(v_i\right) \hat{w}^2_i}|~.
    \end{align*}

Next, one can see that $\sum\limits_{i \in S_1} sign\left(u_i\right) \hat{w}^1_i \sim \Ncal\left(\zero, \frac{k_1}{d}I_\ell\right) $, and $ \sum\limits_{i \in S_2} sign\left(v_i\right) \hat{w}^2_i \sim \Ncal\left(\zero, \frac{k_2}{d}I_\ell\right)$.

Therefore by Lemma C.3 in \citet{melamed2023adversarial} we get that:

\[ \Pr \left[|\inner{\sum\limits_{i \in S_1} sign\left(u_i\right) \hat{w}^1_i , \sum\limits_{i \in S_2} sign\left(v_i\right) \hat{w}^2_i}| \geq \frac{\sqrt{2\ell}}{d}  m\sqrt{k_1}  \right] \leq e^{-\ell/16} + 2e^{-m^2/2 k_2}~.\]

And therefore,
\begin{align*}
    &\Pr \left[|\inner{\Tilde{g}_1, \Tilde{g}_2}|  \geq \frac{\sqrt{2\ell}}{d}    \right]  \\
    =&\Pr \left[\frac{1}{m^2}|\inner{\sum\limits_{i \in S_1} sign\left(u_i\right) \hat{w}^1_i , \sum\limits_{i \in S_2} sign\left(v_i\right) \hat{w}^2_i}| \geq \frac{\sqrt{2\ell}}{d}    \right]  \\
    \leq& \Pr \left[|\inner{\sum\limits_{i \in S_1} sign\left(u_i\right) \hat{w}^1_i , \sum\limits_{i \in S_2} sign\left(v_i\right) \hat{w}^2_i}| \geq \frac{\sqrt{2\ell}}{d} m\sqrt{k_1}   \right]\\
    \leq & e^{-\ell/16} + 2e^{-m^2/2 k_2}\\
    \leq & e^{-\ell/16} + 2e^{-m/2}~.
\end{align*}

As $0 \leq k_1, k_2 \leq m$. 

For the corollary to hold, we note that Lemma C.3 uses Lemma C.1 and the assumption that the norm of the normally distributed vector is not too big. For $w \sim \mathcal{N}(\zero,\sigma^2 I_n)$:
\[
        \Pr \left[\norm{w}^2 \geq 2 \sigma^2 n \right] \leq e^{-\frac{n}{16}}~.
        \] 

Therefore, to conclude that 

\[
|S_C(\Tilde{g}_1, \Tilde{g}_2)|  \leq \Theta\left(\frac{1}{\sqrt{\ell}}\right)~,
\]

One should only note that under the corollary assumption, both $\Tilde{g}_1$ and $\Tilde{g}_2$ having $\Theta(1)$ norm with probability $\geq 1- e^{-\Theta(d)}$.
\end{proof}
\newpage
\section{Additional Information for IMDB Experiment}\label{appen:imdb}
The classifier architecture found in \url{https://www.kaggle.com/code/arunmohan003/sentiment-analysis-using-lstm-pytorch/notebook}, and the dataset taken from \url{https://www.kaggle.com/code/arunmohan003/sentiment-analysis-using-lstm-pytorch/input}. We used the same hyper-parameters as in the cited notebook, trained for 50 epochs.

\subsection{Other Explanation Algorithms}
\begin{figure}[H]
\includegraphics[width=0.8\textwidth]{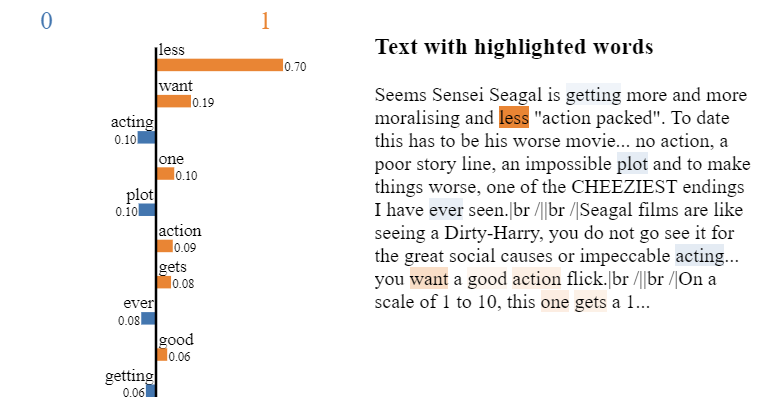}
  \centering
  \caption{LIME textual explanation for IMDB Sentiment Analysis Dataset. One can see for example that "gets" and "getting" are both colored as opposite sentiment related, while both are neutral. In addition, "plot" and "acting" are mistakenly colored as negative sentiment related, as opposed to "action", all quite neutral.}
  \label{img:imdb_lime}
  
\end{figure}

\begin{figure}[H]
\includegraphics[width=0.8\textwidth]{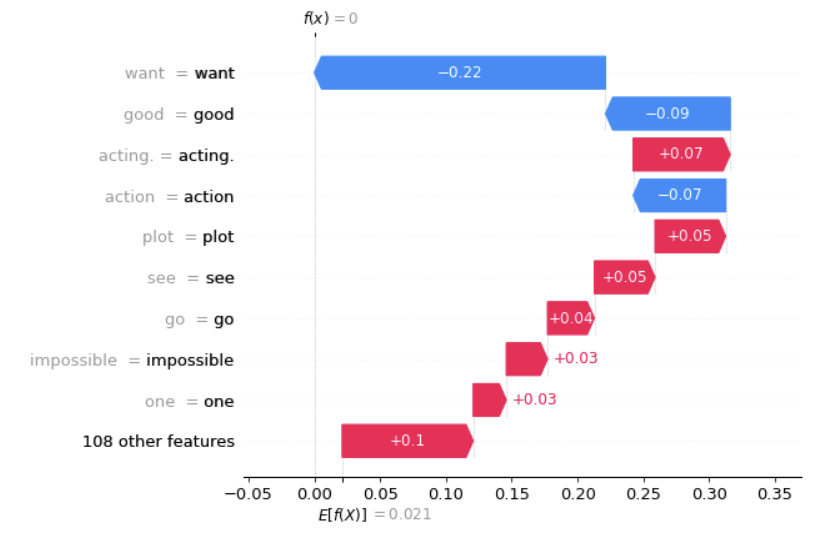}
  \centering
  \caption{SHAP explanation for IMDB Sentiment Analysis Dataset. One can see here too, mostly neutral words picked by the algorithm.}
  \label{img:imdb_shap}
  
\end{figure}

We add two more explanations using the popular LIME and SHAP methods in Figure \ref{img:imdb_lime} and Figure \ref{img:imdb_shap}, respectively. One can see in both examples, that the explanations point out a few of the big-norm words picked also by the classical gradient algorithm chooses words by the norm. One can show an analysis for which, for simple cases, the LIME and SHAP methods are indeed also biased toward the adversarial directions from the input.

\newpage
\section{Additional Information for Security Detection Experiment}\label{appen:Security}
\subsection{Choosing threshold}
\begin{figure}[H]
\includegraphics[width=\textwidth]{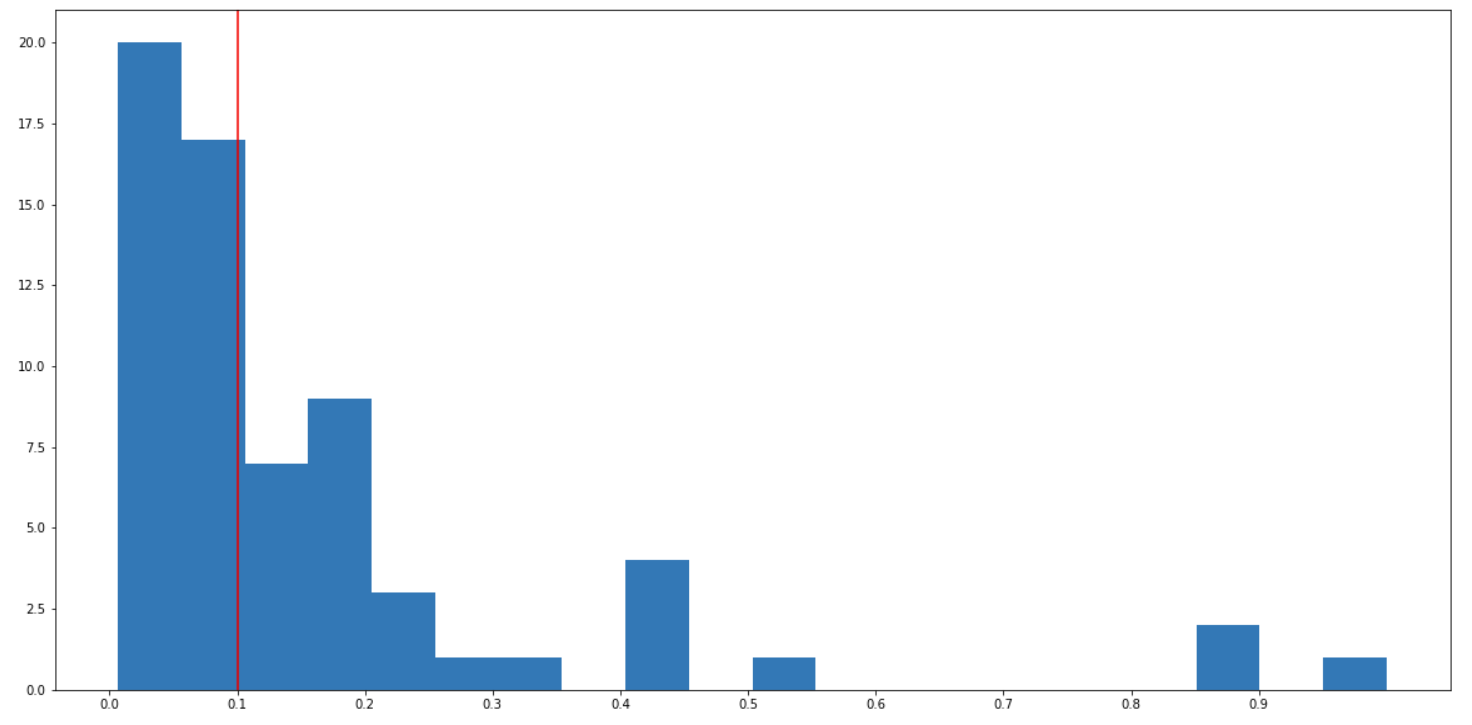}
  \centering
  \caption{Word Gradient Norm for PowerShell Malware Detection Dataset. One can see a clear cut around norm $0.1$.}
  \label{img:powershell_norms}
  
\end{figure}

For this experiment, we used the same threshold $0.1$ for the norms (normalized using $L_{\infty}$ norm for convenience). In Figure \ref{img:powershell_norms} we show that this threshold still seems to separate well the big norms from the smaller ones, as the bars are dramatically shorter right to the $0.1$ threshold. For visualization, the negligible norms (less than $e^{-3}$) are filtered. This short test can be done for each input separately or for the entire dataset together, according to user preferences and application.

\subsection{Other Explanation Algorithms}
\begin{figure}[H]
\includegraphics[width=0.8\textwidth]{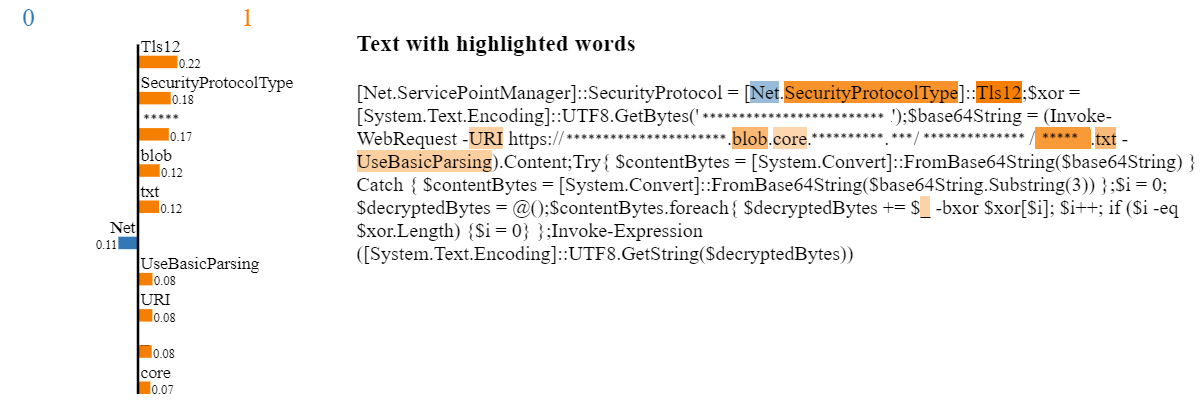}
  \centering
  \caption{LIME textual explanation for PowerShell Malware Detection Dataset. One can see similar word choosing to the big-norm method: choosing of neutral coding phrases and canonical file names shared for all input data.}
  \label{img:powershell_lime}
  
\end{figure}

\begin{figure}[H]
\includegraphics[width=0.8\textwidth]{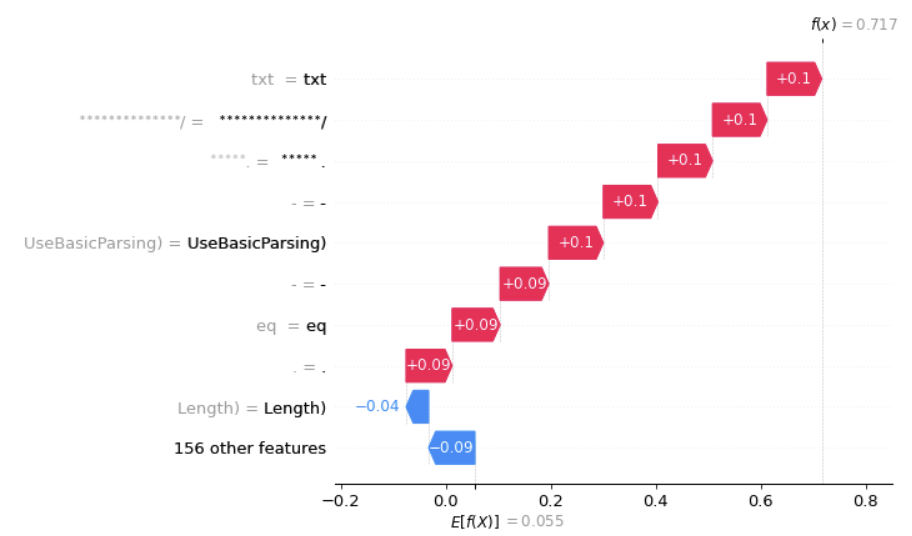}
  \centering
  \caption{SHAP explanation for PowerShell Malware Detection Dataset. One can see again canonical phrases chosen as well as a hyphen, a dot, and neutral words.}
  \label{img:powershell_shap}
  
\end{figure}

Here we add two more explanations by the widely used LIME and SHAP algorithms for text inputs in Figure \ref{img:powershell_lime} and Figure \ref{img:powershell_shap}, respectively. Here too, we can see that the two algorithms picked a word that had a big gradient norm, as seen in Section \ref{sec:security} in Table \ref{tbl_PowerShell} in the max-norm row. We censor the user-sensitive information. In the pre-processing before training, we replace private user information with canonical saved words, here one can see two of them have been mistakenly chosen by the explanation methods.

\end{document}